\documentclass{bmvc2k}
\usepackage{mathrsfs}
\usepackage{amsmath}
\usepackage{amssymb}

\title{UV-Based 3D Hand-Object Reconstruction with Grasp Optimization}

\addauthor{Ziwei Yu}{yuziwei@u.nus.edu}{1}
\addauthor{Linlin Yang\textsuperscript{1, }}{yangll@comp.nus.edu.sg}{2}
\addauthor{You Xie}{you.xie@tum.de}{3}
\addauthor{Ping Chen}{redcping@gmail.com}{4}
\addauthor{Angela Yao}{ayao@comp.nus.edu.sg}{1}

\addinstitution{
National University of Singapore,\\
Singapore
}
\addinstitution{
University of Bonn,\\
Germany
}
\addinstitution{
Technical University of Munich,\\
Germany
}
\addinstitution{
Shanghai Subsidiary of Great Wall Motor Co., Ltd.,\\
China
}
\runninghead{Z. Yu ET AL.}{UV-Based 3D Hand-Object Reconstruction}

\def\ie{\emph{i.e}\bmvaOneDot}

\newcommand{\ho}{{hand-object~}}

\newcommand{\bx}{{\mathbf{x}}}

\newcommand{\bz}{{\mathbf{z}}}

\begin{document}

\maketitle

\begin{abstract}
We propose a novel framework for 3D hand shape reconstruction and hand-object grasp optimization from a single RGB image. The representation of hand-object contact regions is critical for accurate reconstructions. Instead of approximating the contact regions with sparse points, as in previous works, we propose a dense representation in the form of a UV coordinate map. Furthermore, we introduce inference-time optimization to fine-tune the grasp and improve interactions between the hand and the object. Our pipeline increases hand shape reconstruction accuracy and produces a vibrant hand texture. Experiments on datasets such as Ho3D, FreiHAND, and DexYCB reveal that our proposed method outperforms the state-of-the-art.
\end{abstract}

\section{Introduction}
\label{sec:intro}
3D reconstruction of hand-object interactions is critical for 
object grasping in augmented and virtual reality (AR/VR) applications~\cite{huang2020hand,yang2019disentangling,wan2020dual}
We consider the problem of estimating a 3D hand shape when the hand interacts with a known object with a given 6D pose. This set-up lends itself well to AR/VR settings where the hand interacts with a predefined object, perhaps with markers to facilitate the object pose estimation. Such a setting is common, although the majority of previous works~\cite{jiang2021hand,karunratanakul2020grasping,karunratanakul2021skeleton} consider 3D point clouds as input, while we handle the more difficult case of monocular RGB inputs.  Additionally, the previous works~\cite{jiang2021hand,christen2021d,christen2022d,karunratanakul2020grasping,karunratanakul2021skeleton,taheri2020grab} are singularly focused on reconstructing feasible hand-object interactions. They aim to produce hand meshes with minimal penetration to the 3D object without regard for the accuracy of the 3D hand pose. We take on the additional challenge of balancing realistic hand-object interactions with accurate 3D hand poses.  

Representation-wise, previous hand-object 3D reconstruction works ~\cite{boukhayma20193d,khamis2015learning,tzionas2016capturing,zimmermann2019freihand} predominantly with the MANO model~\cite{romero2017embodied}.
MANO is convenient to use, but its accuracy is limited because it cannot represent direct correspondences between the RGB input and the hand surface.
This work considers dense representations and, in particular, focuses on UV coordinate maps. UV maps are ideal as they establish dense correspondences between 3D surfaces and 2D images and work well in representing the 3D human body and face~\cite{feng2018joint,yao2019densebody}. 

\begin{figure*}[t!]
	\centering{
	\includegraphics[width=1.0\linewidth]{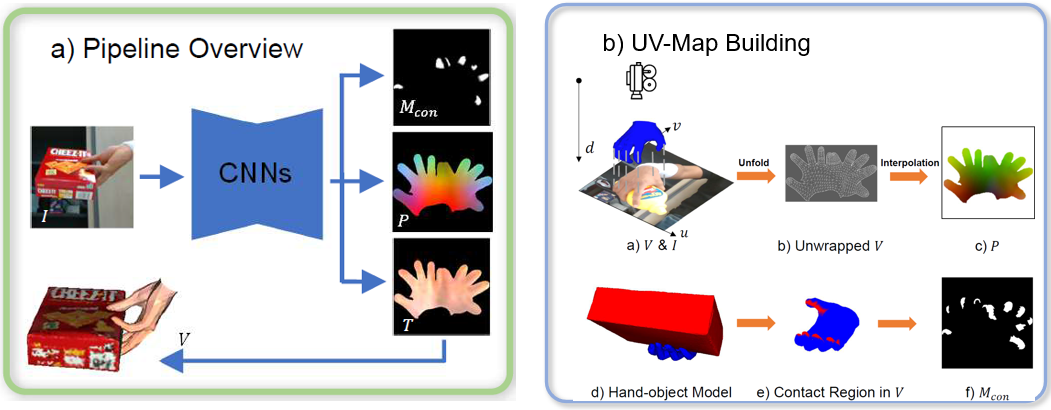}
	\caption{a) We use UV coordinate map $P$ as a dense representation of hand mesh $V$, and train networks to directly generate $P$, contact mask $M_{\text{con}}$ and texture map $T$ from RGB image $I$; we then achieve a 3D reconstruction of the hand-object interaction or just the hand. Our method. b) Generation of the ground truth UV maps (a-c) and contact mask (d-f).
	}
	\label{fig:teaser}}
\end{figure*}

Working with a UV coordinate map has a natural advantage in that we can adopt standard image-based CNN architectures.  The CNN captures the pixels' spatial relationships to aid the 3D modelling while remaining fully convolutional. UV coordinate maps can be easily augmented with additional corresponding information, such as surface texture and regions of object contact. This creates a seamless connection between the 3D hand shape, its appearance, and its interactions with objects in a single representation space (see Fig.~\ref{fig:teaser}a). To that end, we propose an RGB2UV network that simultaneously estimates the hand UV coordinates, hand texture map, and hand contact mask.

The hand contact mask is a key novelty of our work. We propose a binary UV contact mask~(see Fig.~\ref{fig:teaser}b-f) that marks contact regions between the hand surface and the interacting object surfaces. To our knowledge, the contact mask is the first dense representation of hand-object contact. Previous methods~\cite{hasson2019learning,yang2021cpf} work sparsely on a per-vertex or per-point basis and have an obvious efficiency-accuracy trade-off.  The more vertices or points considered, the more accurate the contact modelling and the higher the computational price.  Working in the UV space allows dense contact modelling and improves reconstruction accuracy while remaining efficient.

Normally, accurate hand surface reconstructions in isolation cannot guarantee realistic hand-object interactions; ensuring realistic interactions in vice-versa may result in inaccurate hand reconstructions~\cite{hasson2019learning,li2021artiboost,hasson2020leveraging}.
The main reason is that the hand and object models (MANO~\cite{romero2017embodied},  YCB~\cite{calli2015ycb}) are all rigid.  The rigid assumption is too strong at the contact points, and leads to one mesh penetrating the other~\cite{yang2021cpf,jiang2021hand}, even for ground truth poses. To mitigate these errors, we propose an additional optimization-based refinement procedure to improve the overall grasp. The optimization is performed only during inference and reduces the distances from a hand surface that either penetrates or does not make contact with the object surface. This grasp optimization step significantly improves the feasibility of the hand-object interaction while ensuring accurate hand poses.

Our contributions are summarized as follows: (1) We propose a UV-map-based pipeline for 3D hand reconstruction that simultaneously estimates UV coordinates, hand texture, and contact maps in UV space. (2) Our model is the first to explore dense representations to capture contact regions for hand-object interaction. (3) Our grasp optimization refinement procedure yields more realistic and accurate 3D reconstructions of hand-object interactions. (4) Our model achieves state-of-the-art performance on 
FreiHAND, Ho3D and DexYCB hand-object reconstruction benchmarks.

\section{Related Works}

\textbf{Face, Body and Hand Surface Reconstruction.} 
A popular way to represent human 3D surfaces is via a 3D mesh. Previous works estimate mesh vertices either directly~\cite{ge20193d,choi2020pose2mesh,wan2020dual,moon2020i2l,ranjan2018generating,lee2020uncertainty}, or indirectly~\cite{boukhayma20193d,zimmermann2019freihand,kolotouros2019learning} through a parametric model like SMPL~\cite{loper2015smpl} for the body, 3DMM~\cite{blanz1999morphable} for the face, and MANO~\cite{romero2017embodied} for the hand. Parametric models are convenient but cannot provide direct correspondences between the input images and the 3D surface. As such, we learn a dense surface as a UV coordinate map.  We are inspired by the success of recent UV works for the human body~\cite{guler2018densepose,zeng20203d}, face~\cite{deng2018uv,lee2020styleuv,kang2021competitive}, and hand~\cite{chen2021i2uv,wan2020dual}. To the best of our knowledge, our work is the first to explore the use of UV coordinate maps in modelling hand-object interactions and capturing the contact regions.

\textbf{Texture Learning on Hand Surfaces.} Shading and texture cues are often leveraged to solve the self-occlusion problem for hand tracking~\cite{de2011model,de2008model}. Recently,~\cite{chen2021model} verified that texture modelling helps with the self-supervised learning of shape models. Instead of relying on a parametric texture model~\cite{qian2020parametric}, we apply a UV texture map. UV texture maps are commonly used to model the surface texture of the face~\cite{deng2018uv,lee2020styleuv,kang2021competitive} and, more recently, the human body~\cite{zeng20203d}. We naturally extend UV texture maps for hand-object interactions.

\textbf{Hand-Object Interactions.} Various contact losses have been proposed~\cite{hasson2019learning,yang2021cpf,hasson2021towards} to ensure feasible hand-object interactions.  Examples such as repulsion loss and attraction loss ensure that the hand makes contact with the object but avoids actual surface penetration. Contact regions, however, are modeled sparsely. Contact points are sparse and selected either statistically~\cite{hasson2020leveraging}, or based on hand vertices closest to the object~\cite{li2021artiboost}. 
These methods have an accuracy-efficiency trade-off hinging on the number of considered vertices. In contrast, our UV contact mask establishes a dense correspondence between the contact regions and the local image features.  It is both accurate and efficient.  
Our work is the first to explore a joint position, texture and contact-region UV representation for hand-object interactions. 

\section{Method}
\begin{figure*}[!t]
	\centering{
	\includegraphics[width=1.0\linewidth]{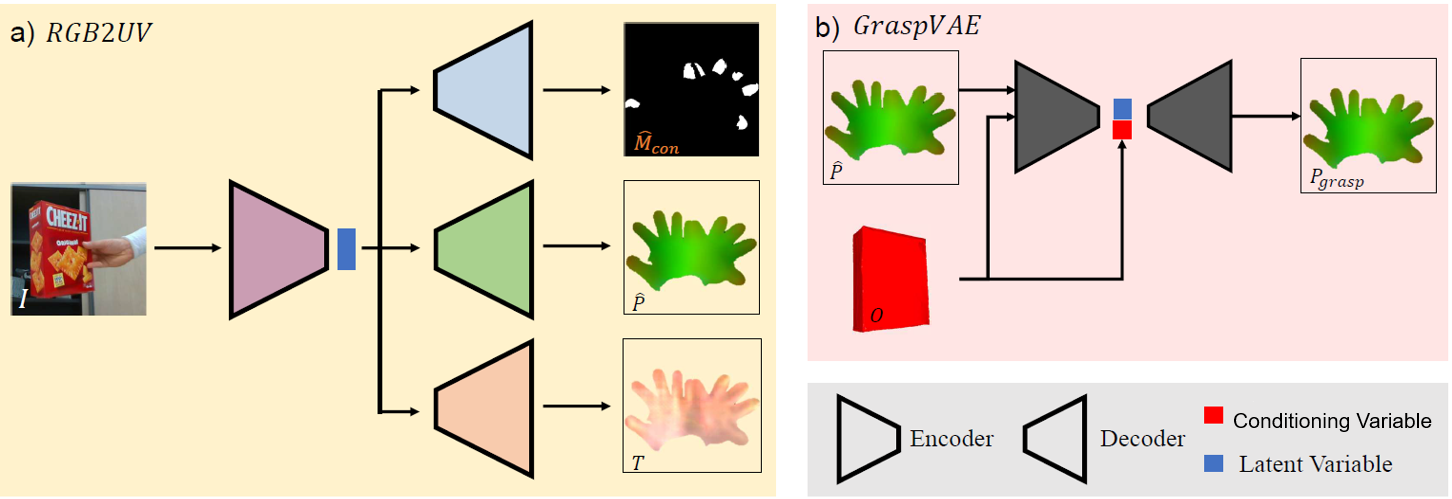}
	\caption{Overview of a) RGB2UV and b) GraspVAE. RGB2UV takes as input an image and outputs the hand contact mask, UV coordinate map, and texture map. GraspVAE further refines the predicted UV coordinate map by conditioning on the interacting object. }
    \label{fig:pipeline}}
\end{figure*}

Our method has two components: the RGB2UV network and grasp optimization (see Fig.~\ref{fig:pipeline}). The RGB2UV network (Sec.~\ref{subsec:uv} and~\ref{subsec:architecture}) takes as input an RGB image of a hand-object interaction and outputs a hand UV mask and contact mask. The grasp 
optimization (Sec.~\ref{subsec:grasp}) adjusts the mesh vertices to refine the hand grasp and reduces penetrations.
\subsection{Ground Truth UV, Contact Mask \& Texture}\label{subsec:uv}
\textbf{UV Coordinate Map.}
Suppose we are given a 3D hand surface in the form of a 3D mesh $V$ and an accompanying RGB image $I$ (see Fig.~\ref{fig:teaser} a). The hand mesh $V (\bx_V)\!\in\!\mathbb{R}^{778 \times 3}$ stores the 3D position $\bx_V=(x,y,z) \in \mathbb{R}^{3}$ of the 778 vertices in the camera coordinate space.  The image $I(\bx_{I}) \in \mathbb{R}^{H_I \times W_I \times 3}$ records the RGB values of each pixel $\bx_I=(u,v) \in \mathbb{R}^{H_I \times W_I}$ in the image plane. $I$ can be considered a projection of $V$ into 2D space, with the mapping 
\begin{equation}
(u,v,d) = \mathcal{T}((x,y,z),c),
\label{eq:projection}
\end{equation}
where $\mathcal{T}(\bx_V,c)$ is a transformation that can project a vertex at $(x,y,z)$ into $(u,v,d)$ in the image plane with camera parameters $c$. Here, $d$ denotes the distance from the camera to the image plane. Because $I$ is on the image plane, it only reveals the vertices closest to the camera; 
we thus consider the vertical distance $d$ from the camera to the vertex (see Fig.~\ref{fig:teaser} b-a), and represent each vertex uniquely with $(u,v,d)$.

To represent all the hand vertices, even those not seen in $I$, in one 2D plane, we unwrap the MANO mesh with MANO's provided UV unfolding template (see Fig.~\ref{fig:teaser} b-b)~\cite{romero2017embodied}) to arrive at a UV coordinate map $P$ with UV coordinates $\bx_P \in \mathbb{R}^{H_P \times W_P \times 3}$, where
\begin{equation}\label{eq:P2UVD}
P(\bx_P) = (u,v,d).
\end{equation}
Mapping only the vertices $V$ onto the UV map results in a sparse coordinate map. To generate a dense UV representation $P$ (Fig. \ref{fig:teaser} b-c), we 
interpolate values on each triangular mesh face from neighbouring vertices. 
Given $(u,v,d)$, the corresponding 3D coordinate is given by the $\mathcal{T}^{-1}$ is the inverse transformation of $\mathcal{T}$ from Eq.~\ref{eq:P2UVD}:
\begin{equation}
(x,y,z) = \mathcal{T}^{-1}((u,v,d),c).
\label{eq:UV2XY}
\end{equation}

\noindent \textbf{Contact Mask.} Based on $P$, we propose a binary contact mask $M_{\text{con}}\in \mathbb{R}^{H_P \times W_P}$ to indicate the contact region in hand-object interactions. Like~\cite{hampali2020honnotate}, contact vertices are defined as the vertices within 4 mm to the grasped object.  We first locate contact vertices $\bx_V$ (Fig. \ref{fig:teaser} b-e) and their corresponding points $\bx_P$ in $P$ with Eq. \ref{eq:projection} and Eq. \ref{eq:P2UVD}. 
 The contact mask is defined as: 
\begin{equation}
M_{\text{con}}(\bx_P) = \left\{\begin{matrix}
0,\qquad {\text{if}~dist(\textbf{v}, V_O) > 4~mm;}\\ 
\hspace{3mm}1,\qquad \text{if}~dist(\textbf{v}, V_O) <= 4~mm.
\end{matrix}\right.
\end{equation}

\noindent Similar to $P$, direct mapping from $V$ to $M_{\text{con}}$ generates a sparse representation. For every triangular plane in $V$, we interpolate all points inside the plane as a contact region if the three vertices are contact vertices to achieve a dense representation of $M_{\text{con}}$ (see Fig. \ref{fig:teaser} b-f).

\noindent \textbf{Texture.} The UV coordinates $P$ also conveniently establish a correspondence between the mesh vertices and texture information. The texture map $T \in \mathbb{R}^{H_P \times W_P \times 3}$ gives information on hand appearance.  For every point $\bx_P$ in $T$, similar to Eq.~\ref{eq:P2UVD}, we obtain the texture map:
$T(\bx_P)=I(u,v)$.
With Eq.~\ref{eq:UV2XY}, the texture of the vertices in the hand mesh can be tracked in $T$.

\subsection{UV Estimation from an RGB Image}\label{subsec:architecture}
The RGB2UV network, $(\hat{P},\hat{M}_{\text{con}},T)=RGB2UV(I)$, estimates the UV coordinate map $\hat{P}$, contact mask $\hat{M}_{\text{con}}$, {and texture map $T$} from an input image $I$.  With prepared data tuples $(I,P,M_{\text{con}})$, as described in Sec.~\ref{subsec:uv}, the network can be trained in a supervised way. Similar to previous works~\cite{chen2021i2uv,zeng20203d,yao2019densebody}, we use a U-Net style encoder-decoder with a symmetric ResNet50 and skip connections between the corresponding encoder and decoder layers. 

Our network features {three} decoders: a hand UV coordinate map decoder, a contact mask decoder, and a UV texture decoder (see Fig~\ref{fig:pipeline} a). The encoder and three decoders are trained jointly with five losses:
\begin{equation}
\begin{aligned}
L_{\text{RGB2UV}} =  L_{P} +  L_{\text{grad}} + \lambda_1 L_{M_{\text{con}}}+  L_{V}+\lambda_2 L_{\text{texture}} \\
\end{aligned}
\end{equation}

\noindent \textbf{UV Coordinate Map Decoder.} This decoder is learned with the hand UV coordinate map loss $L_{P}$, hand UV gradient loss $L_{\text{grad}}$. These losses are L1 terms between the ground truth and the predicted UV coordinate maps ${P}$ and $\hat{P}$, and their respective gradients $\nabla P$ and $\nabla {\hat{P}}$:
\begin{equation}
L_{\text{P}} = |P - \hat{P}|\cdot M \qquad \text{and} \qquad L_{\text{grad}} = | \nabla P - \nabla \hat{P}| \cdot M,
\end{equation}
where $M$ is a mask in the UV space indicating the entire hand region. $M$ is applied to constrain the consideration of the UV coordinate map and their gradients to only valid regions on the hand's surface in $\hat{P}$ and $\nabla{\hat{P}}$.

\noindent \textbf{Contact Mask Decoder.}  This decoder is learned with the L1 contact mask loss $L_{M_{\text{con}}}$ between the ground truth and predicted contact mask $M_{\text{con}}$ and $\hat{M}_{\text{con}}$ respectively: $ L_{M_{\text{con}}} = |M_{\text{con}} - \hat{M}_{\text{con}}|$. To ensure correctness in the hand pose represented by the coordinate map and contact mask decoders, we add an additional mesh vertex loss $L_{V}$. The hand mesh $V$ implicitly represents the pose of the hand, so a loss between ground truth $V$ and estimated $\hat{V}$ improves the accuracy of the estimated pose. The pose loss $L_{V}$ is defined as: $L_{V} = |V - \hat{V}|,$ where $V$ represents ground truth hand mesh vertices and $\hat{V}$ is sampled from $\hat{P}$.

\noindent \textbf{Texture Decoder.} As there is no ground truth for UV texture, we apply {self-supervised} training to estimate the UV texture map. 
With given camera parameters $c$, the ground truth mesh vertices $V$ can be projected into the $uv$ plane to obtain a view-specific hand silhouette $S_{\text{gt}}$ via a differentiable renderer\footnote{We use the built-in renderer from the \texttt{pytorch3D} library~\cite{ravi2020pytorch3d}.}. The hand region in RGB image $I$ can be isolated with $I \cdot S_{\text{gt}}$ by applying $S_{\text{gt}}$ as a mask. On the other hand, we also render the hand image $I_{re}$ with the estimated UV coordinate and texture map, \ie~$\hat{P}$ and $T$. The texture decoder can then be trained by minimizing the distance between $I_{re}$ and $I \cdot S_{\text{gt}}$. We apply a photometric consistency loss $L_{\text{texture}}$~\cite{chen2021model} for training, which consists of an RGB pixel loss term $L_\text{pixel}$ and a structure similarity loss $L_\text{SSIM}$ based on the structural similarity index~\cite{brunet2011mathematical} $\text{SSIM}$:

\begin{equation}
L_{\text{texture}} =   L_{\text{pixel}} + \lambda_3 L_{\text{SSIM}}, \quad L_{\text{pixel}} = |I \cdot S_{\text{gt}} - I_\text{re}|, {\text{and}}\quad L_{\text{SSIM}} = 1 - \text{SSIM}(I \cdot S_{\text{gt}}, I_\text{re}).
\end{equation}

\subsection{Grasp Optimization} \label{subsec:grasp}
The RGB2UV network is standalone and does not take any object information into consideration. Even though the ground truth 6D object pose is provided in our setup, this does not ensure feasible hand grasps or naturalistic \ho interactions. Therefore, we introduce a grasp optimization step for further refinement. Instead of refining the 3D hand mesh directly, which is very high-dimensional with 778 vertices, we work in a latent variable space 
to reduce the dimensionality of the optimization. Specifically, we learn a conditional VAE with an encoder $En(\cdot)$ and a decoder $De(\cdot)$, which we name GraspVAE. 

During inference, the estimated UV coordinate map $\hat{P}$ is encoded into a latent variable $\bz$ with GraspVAE's $En(\cdot)$. The latent $\bz$ is refined to $\bz^*$ through optimization to minimize the penetrations between the object mesh and the estimated hand mesh. Finally, $\bz^*$ is decoded with $De(\cdot)$ into a UV coordinate map and converted into a hand mesh.  Note the optimization happens during inference only, whereas $En(\cdot)$ and $De(\cdot)$ are learned during training. 

\noindent \textbf{GraspVAE} is a conditional VAE with a ResNet18 encoder and three transposed convolution layers as the decoder.  It is conditioned on the object vertices $O$ (see Fig.~\ref{fig:pipeline} b) to estimate a refined UV coordinate map $P_\text{grasp}$ from the estimated $\hat{P}$ from the RGB2UV network, \ie~$P_\text{grasp} = \text{GraspVAE}(\hat{P} | O)$.  GraspVAE is trained with the combined loss
\begin{equation}\label{eq:VAE}
L_{\text{Grasp}} =  L_{P} +  L_{\text{grad}} +  L_{V} + \lambda_{4} L_{\text{KL}}+  L_{\text{pene}}, \quad \text{where} \quad
L_{\text{pene}} =\frac{1}{|V_{in}^o|}\sum_{\bx_V \in V_{in}^o} \text{dist}(\textbf{v},V_O).
\end{equation}
\noindent Here, $L_{\text{KL}} = KL(q(\bz|(\hat{P},O))||p)$ is the standard Kullback-Leibler divergence loss used in VAE models, where $\bz$ represents the latent variables encoded from input $(\hat{P},O)$. The term $p= \mathcal{N}(0, E)$ denotes a Gaussian prior where $E$ is an {identity matrix}. The term $q(\bz|(\hat{P},O))$ is the distribution of $\bz$, while the $\text{dist}(\cdot, \cdot)$ function estimates the closest distance between a hand vertex $\bx_V$ and object vertices $O$. The penetration loss $L_{\text{pene}}$ is applied to penalize hand vertices $\bx_V \in V_{in}^O$, where $ V_{in}^O$ represents the set of hand vertices inside the object. Existing works~\cite{hasson2019learning} identify $V_{in}^O$ from the entire hand mesh, which is computationally inefficient. Our contact mask restricts the search area and decreases the processing time. For more details on training a conditional VAE, we refer the reader to~\cite{cvae}.

\noindent \textbf{Latent-Space Optimization.} During inference, the learned GraspVAE model is used to estimate a refined UV map.  Specifically, we encode the given object vertices $O$ and estimated $\hat{P}$, \ie $\bz_o = \text{En}(\hat{P},O)$ and solve for an optimized $\bz^*$, with $\bz_o$ as initialization: 
\begin{equation}
     \bz^{*} = {arg\,\min_{\bz}} \; L_{\text{KL}} +  L_{\text{pene}}.
\end{equation}
   
\noindent The hand UV map is generated with $P^*=De(\bz^{*}\bigoplus O)$, where $\bigoplus$ denotes a concatenation operation. The refined hand mesh $V^*$ can then be generated from $P^*$ with Eqs.~\ref{eq:P2UVD} and \ref{eq:UV2XY}. More details are given in the Supplementary.

\noindent \textbf{Hand Only VAE.}
To further highlight the strengths of our grasp optimization, 
we introduce a variant of GraspVAE named Hand Only VAE to refine the hand shape accuracy with. Similar to GraspVAE, the encoder of Hand Only VAE is a ResNet18 network and the decoder contains three transposed convolution layers. Hand Only VAE is trained with $L_{\text{Grasp}}$ in Eq. \ref{eq:VAE}, but without $L_{\text{pene}}$. 
Thus, the role of $L_{\text{pene}}$ in grasp optimization can be visualized from differences between results of Hand Only VAE and GraspVAE.

\section{Experiments}
\noindent\textbf{Implementation Details.}
The Adam optimizer is applied to train all networks over 80 epochs with a batch size of 64. 
We start with an initial learning rate of $10^{-4}$ for all training and lower it by a factor of 10 at the 20th, 40th and 60th epochs.
After GraspVAE is trained,  we use the learning rate of $10^{-6}$ to update the latent variables until the loss difference is lower than $10^{-6}$ for grasp optimization. The parameters are set empirically all to 1, except for 
$\lambda_{1,2,3} = 10$, $\lambda_{4} = 0.001$.  The VAE latent space is 128 dimensions. 

\noindent\textbf{Datasets.} We evaluate on RGB-based \ho benchmarks HO3D~\cite{hampali2020honnotate,hampali2021ho}, DexYCB~\cite{chao2021dexycb}, and FreiHAND~\cite{zimmermann2019freihand}. 
We compare HO3D with state-of-the-art for both v2 and v3
and report our results through their leaderboard (v2\footnote{https://competitions.codalab.org/competitions/22485\#results }, v3\footnote{https://competitions.codalab.org/competitions/33267\#results }). 
For DexYCB dataset, we use the official ``S0'' split.
Since {FreiHAND} does not provide object models and object annotations, we only evaluate the hand shape reconstruction on our RGB2UV network and our Hand Only pipeline. Results are reported through their 
leaderboard\footnote{https://competitions.codalab.org/competitions/21238\#results}.  

\noindent\textbf{Metrics.} For evaluating the hand pose and shape accuracy, we use the mean-per-joint-position-error (MPJPE) (cm) for 3D joints and mean-per-vertex-position-error (MPVPE) (cm) for mesh vertices.  For evaluating the hand-object interaction, we measure the penetration depth (PD) (mm), solid intersection volume (SIV) ($cm^3$)~\cite{yang2021cpf} and  Simulation Displacement (SD) (cm)~\cite{hasson2019learning}.  PD 
is based on the maximum distance of all hand vertices inside the object with respect to their closet object vertices. SIV is defined as the total voxel volume of hand vertices inside the object after converting the object model into $80^3$ voxels. SD measures grasp stability in a simulation space that the hand is fixed and the grasped object is subjected to gravity.  

\begin{table*}[!tb]
\centering
\scriptsize
\scalebox{0.97}
{
\begin{tabular}{p{1.5cm}|p{0.25cm}p{0.25cm}|p{0.25cm}p{0.25cm}p{0.25cm}p{0.25cm}p{0.25cm}|p{0.25cm}p{0.25cm}p{0.25cm}p{0.25cm}p{0.25cm}|p{0.2cm}p{0.2cm}p{0.2cm}p{0.2cm}p{0.2cm}}
\hline
\multicolumn{1}{l|}{Dataset} &\multicolumn{2}{c|}{FreiHand}  &\multicolumn{5}{c|}{Ho3D V2}
&\multicolumn{5}{c|}{Ho3D V3} &\multicolumn{5}{c}{DexYCB} \\ 
\cline{1-18} 
Method &\tiny{MPJPE}&\tiny{MPVPE} &\tiny{MPJPE} &\tiny{MPVPE} &\tiny{PD} &\tiny{SIV} &\tiny{SD}  &\tiny{MPJPE} &\tiny{MPVPE} &\tiny{PD} &\tiny{SIV} &\tiny{SD} &\tiny{MPJPE} &\tiny{MPVPE} &\tiny{PD} &\tiny{SIV} &\tiny{SD} 
\\\hline
{Hampali}\cite{hampali2020honnotate} &- &- &1.07 &1.06 &12.34  &17.28 &4.02 &- &- &- &-  &- &- &- & - &- & -\\
{Hasson}\cite{hasson2019learning} &1.33 &1.33  &1.10 &1.12 &- &- &- &- &- &- &- &- &- &- &- &- &-  \\  
{Hasson}\cite{hasson2020leveraging} &1.33 &1.33 &1.14 &1.09  &18.44 &14.35 &4.10 &- &- &- &- &- &- &- &- &- &-  \\ 
{\tiny{GraspField}} \cite{karunratanakul2020grasping}&- &- &1.38 &1.36  &14.61 &14.92 &\textbf{3.29} &- &- &- &- &- &- &- &- &- &-\\
Li~\cite{li2021artiboost} &- &- &1.13  &1.10 &11.32   &14.67 &3.94 &\underline{1.08} &\underline{1.09}  &16.78 &\underline{10.87} &3.90 &1.28 &1.33 &9.47 &11.02  &3.64  \\ 
Chen~\cite{chen2021i2uv}  &\underline{0.72} &\underline{0.74} &\textbf{0.99} &\textbf{1.01} &10.25  &15.38 &4.05 &1.25 &1.24  &17.87 &12.56 &4.02 &\underline{1.23} &\underline{1.13} &8.64 &12.38 &3.36   \\
Dataset GT & - &- &- &- &-  &- &- &-  &- &- &- &- &0 &0 &4.58 &7.16   &1.48  \\  \hline
MANO CNN  &1.10 &1.09 &1.30 &1.30 &17.42 &14.2 &4.33 &1.38 &1.37   &18.65 &14.78 &4.36 &1.39 &1.40  &16.24 &15.38 &4.08\\  
MANO Fit &1.37 &1.37  &1.58 &1.61  &21.37 &18.0 &4.82 &1.66 &1.65 &22.04 &15.60 &4.65 &1.49 &1.43  &18.56 &17.60 &4.86 \\
RGB2UV &0.75 &0.78 &1.08 &1.07 &11.33 &17.24 &4.24 &1.22 &1.23 &16.72 &13.50 &4.27 &1.20 &1.19 &7.03 &11.02 &3.28\\
Hand Only &\textbf{0.71} &\textbf{0.73}  &\underline{1.04} &\underline{1.04} &\underline{9.68} &\underline{14.1} &3.92 &\textbf{1.08} &\textbf{1.04} &\underline{13.07} &11.77 &\underline{3.88} &\textbf{1.09} &\textbf{1.02} &\underline{6.44} &\underline{9.32} &\underline{2.98} \\
Hand+Object  &- &- &1.25 &1.33 &\textbf{7.66} &\textbf{10.4} &\textbf{3.22} &1.28 &1.26 &\textbf{9.67} &\textbf{8.66} &\textbf{3.01} &1.29 &1.27 &\textbf{5.05} &\textbf{7.11}  &\textbf{2.64} \\ \hline
\end{tabular}}
\caption{{Comparison with state-of-the-art methods. 
\textbf{Best} and \underline{Second-best} Scores. Our Hand Only achieves the best holistic performance across all comparisons. {Note that 
the PD and SIV of DexYCB ground truth data are non-zero due to the rigid modeling of both the hand and the object.  Combining the two inevitably results in some penetration, and this serves as a lower bound in achievable results.}}}
\label{tab:all_results}
\end{table*}

\subsection{Comparison with the State-of-the-Art}
\paragraph{Quantitative Results.}
Comparison with state-of-the-art results of~\cite{chen2021i2uv,li2021artiboost,hasson2019learning,hasson2020leveraging, hampali2020honnotate} in Table \ref{tab:all_results} are based on their released source code and default parameters.
Considering only the hand pose and shape accuracy, our Hand Only pipeline obtains the lowest MPJPE and MPVPE on Ho3D V3 and DexYCB.
Especially on the DexYCB dataset, compared with the latest works, our method reduces the pose error MPJPE by 10\%. For hand-object interaction, our Hand-Object pipeline also achieves the lowest PD, SIV and SD for Ho3D V2 and DexYCB. Comparing the  results for the DexYCB dataset, our grasp results verify the effectiveness of our pipeline.

\begin{figure}[tb!]
	\centering{
	\includegraphics[width=1.0\linewidth]{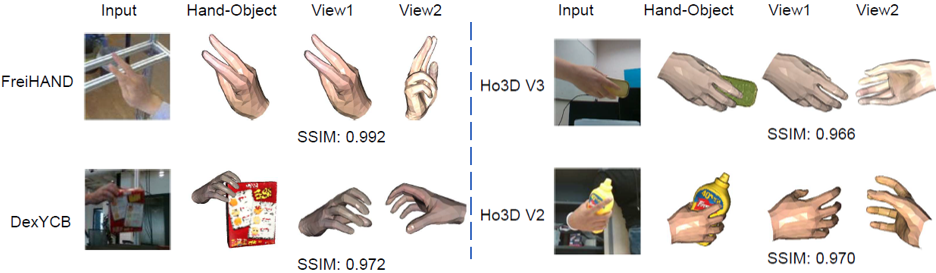}
	\caption{Surface Results: predicted 3D hand surfaces with hand shape. For each quartet, from left to right the columns correspond to RGB input, hand surface with ground truth object, hand surface in camera view, and hand surface in a different view. {Besides, similarities between projected hand $I_{re}$ and its ground truth are measured with the SSIM matrix. High SSIM values indicate that our texture decoder can generate reliable textures.}  }
	\label{fig:surface}
	}
\end{figure}

\begin{figure}[tb!]
	\centering{
	\includegraphics[width=0.8\linewidth]{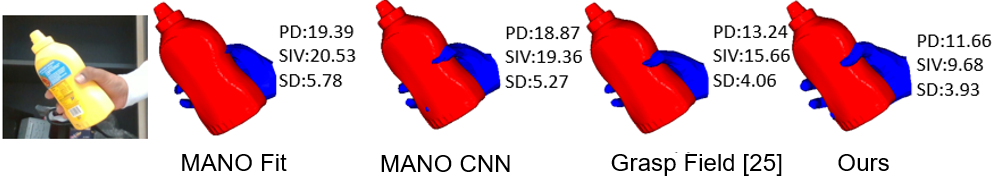}
	\caption{{Grasp Results: Comparison of hand-object reconstructions with state-of-the-art. Our method yields a more feasible hand with the lowest PD, SIV and SD.}
	}
	\label{fig:grap_quality}
	}
\end{figure}

\noindent\textbf{Qualitative Results.}
Visualizations of our hand surfaces in Fig.~\ref{fig:surface} verify our texture learning and grasp optimization effectiveness.Additionally, Fig.~\ref{fig:grap_quality} compares our method to state-of-the-art, demonstrating our improved grasp feasibility. More qualitative results can be found in the supplementary material.

\noindent\textbf{Texture Map Results.}
Visualizations of our results are all rendered with textures from our texture decoder. Fig.~\ref{fig:surface} also reports appearance similarity between the projected hand $I_{re}$ and its ground truth via SSIM (higher is better).
Our model can even estimate the texture for unobserved vertices, \ie the results of view 2, highlighting our model's ability to generalize appearance. 
We further compare our hand texture reconstruction results with~\cite{chen2021model} in the supplementary material and show that our reconstruction results are better.

\begin{figure}[htb]
	\centering{
	\includegraphics[width=0.8\linewidth]{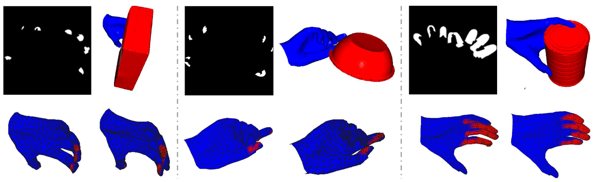}
	\caption{Contact mask results. 
 For each sample, top-left: our predicted contact mask; top-right: our predicted hand surface with ground truth objects; bottom-left: our contact regions from the predicted contact mask; bottom-right: ground truth contact regions.}
	\label{fig:contact_region}}
\end{figure}

\noindent\textbf{Contact Mask Results.}
Fig.~\ref{fig:contact_region} shows predicted contact masks for DexYCB. 
Estimated contact regions are similar to the ground truth.
Our generated contact masks $\hat{M}_\text{con}$ has  
an average Intersection over Union (IoU) of $72.09\%$. 
Efficiency-wise, our contact mask is $5$ times faster than point-based methods~\cite{yang2021cpf,hasson2019learning}. Detailed experimental results are given in the supplementary material.

\subsection{Ablation Study}
{\textbf{MANO Baselines.}
We also compare with two simple MANO baselines in Table~\ref{tab:all_results}.  
MANO CNN is a pipeline that regresses the MANO parameters through the differentiable MANO layer~\cite{zimmermann2019freihand,taheri2020grab}. 
MANO Fit uses inverse-kinematics method to fit MANO parameters from our estimated hand vertices. We outperform these baselines; especially on FreiHAND, our method achieves significant improvements (40\% over MANO CNN and MANO Fit).  }

\begin{figure}[tb!]
	\centering{
	\includegraphics[width=0.8\linewidth]{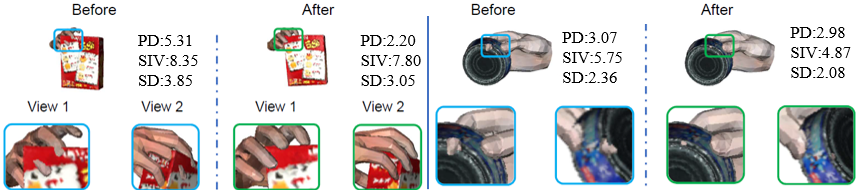}
	\caption{Grasp Optimization: Predicted hand surfaces with ground truth objects before and after grasp optimization. Differences are highlighted in the boxes. Our grasp optimization significantly reduces PD, SIV and SD, and yields more feasible hand-object interactions.}
	\label{fig:visualization}
	}
\end{figure}

\noindent\textbf{GraspVAE.}
The impact of GraspVAE is shown in Table \ref{tab:all_results} (Hand-Only vs Hand-Object).
Grasp optimization worsens MPJPE and MPVPE but improves PD, SIV and SD because the optimization limits penetration between the hand and object at the expense of less accurate hand pose and shape.
Fig.~\ref{fig:visualization} shows examples before and after optimization. Optimization reduces the penetration and this is verified with 
lower PD, SIV and SD. Furthermore, compared with RGB2UV pipeline, using the Hand Only VAE refinement reduces pose errors by nearly 10\%. This reveals the efficiency
of our proposed Hand Only VAE and emphasizes the importance of refinement for predicting UV coordinate maps.

\subsection{Limitations}
\begin{figure}[htb]
	\centering{
	\includegraphics[width=0.8\linewidth]{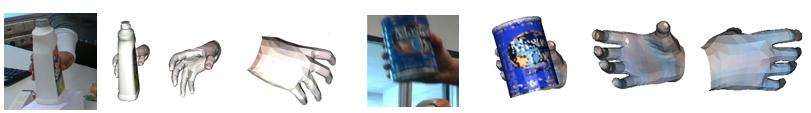}
 	\caption{Failure cases.
  We are limited by only using a single view for texture rendering, 
  and object textures may spill over onto the hand, \ie white and blue hand surfaces from the object.
  }
	\label{fig:limitation}}
\end{figure}

In our RGB2UV pipeline, we used a silhouette of the hand projection $S_{\text{gt}}$ to isolate the hand in the RGB image $I$. If directly use provided object information, the object can be removed by using an object silhouette but it would break the spatial correlation of hand-object interaction. Therefore, without removing hand object may result in spilling over into the hand texture results; this is especially the case when the area of the hand in the image is very small, or there is significant occlusion from the object (see Fig.~\ref{fig:limitation}).
These limitations can be improved by considering multi-view information, which would exclude the extreme viewpoints and enable a feasible hand reconstruction to be obtained.

\section{Conclusion}
This work proposes a hand surface framework for estimating \ho interaction from RGB images. We explored UV coordinate maps for hand-object surface modelling and designed the first dense representation to model contact regions. Additionally, we introduce grasp optimization to improve the feasibility of the hand UV coordinate map. Experimental results show that our proposed method outperforms existing 3D \ho interaction methods.

\section{Acknowledgments}
This research is supported by the Ministry of Education, Singapore, under its MOE Academic Research Fund Tier 2 (STEM RIE2025 MOE-T2EP20220-0015).
\bibliography{bmvcarxiv}

\begin{thebibliography}{72}
\providecommand{\natexlab}[1]{#1}
\providecommand{\url}[1]{\texttt{#1}}
\expandafter\ifx\csname urlstyle\endcsname\relax
  \providecommand{\doi}[1]{doi: #1}\else
  \providecommand{\doi}{doi: \begingroup \urlstyle{rm}\Url}\fi

\bibitem[Blanz and Vetter(1999)]{blanz1999morphable}
Volker Blanz and Thomas Vetter.
\newblock A morphable model for the synthesis of 3d faces.
\newblock In \emph{Proceedings of the 26th annual conference on Computer
  graphics and interactive techniques}, pages 187--194, 1999.

\bibitem[Boukhayma et~al.(2019)Boukhayma, Bem, and Torr]{boukhayma20193d}
Adnane Boukhayma, Rodrigo~de Bem, and Philip~HS Torr.
\newblock 3d hand shape and pose from images in the wild.
\newblock In \emph{CVPR}, 2019.

\bibitem[Brahmbhatt et~al.(2020)Brahmbhatt, Tang, Twigg, Kemp, and
  Hays]{brahmbhatt2020contactpose}
Samarth Brahmbhatt, Chengcheng Tang, Christopher~D Twigg, Charles~C Kemp, and
  James Hays.
\newblock Contactpose: A dataset of grasps with object contact and hand pose.
\newblock In \emph{ECCV}, 2020.

\bibitem[Brunet et~al.(2011)Brunet, Vrscay, and Wang]{brunet2011mathematical}
Dominique Brunet, Edward~R Vrscay, and Zhou Wang.
\newblock On the mathematical properties of the structural similarity index.
\newblock \emph{TIP}, 21\penalty0 (4):\penalty0 1488--1499, 2011.

\bibitem[Cai et~al.(2018)Cai, Ge, Cai, and Yuan]{cai2018weakly}
Yujun Cai, Liuhao Ge, Jianfei Cai, and Junsong Yuan.
\newblock Weakly-supervised 3d hand pose estimation from monocular rgb images.
\newblock In \emph{ECCV}, 2018.

\bibitem[Calli et~al.(2015)Calli, Singh, Walsman, Srinivasa, Abbeel, and
  Dollar]{calli2015ycb}
Berk Calli, Arjun Singh, Aaron Walsman, Siddhartha Srinivasa, Pieter Abbeel,
  and Aaron~M Dollar.
\newblock The ycb object and model set: Towards common benchmarks for
  manipulation research.
\newblock In \emph{ICAR}, 2015.

\bibitem[Cao et~al.(2021)Cao, Radosavovic, Kanazawa, and
  Malik]{cao2021reconstructing}
Zhe Cao, Ilija Radosavovic, Angjoo Kanazawa, and Jitendra Malik.
\newblock Reconstructing hand-object interactions in the wild.
\newblock In \emph{ICCV}, pages 12417--12426, 2021.

\bibitem[Chao et~al.(2021)Chao, Yang, Xiang, Molchanov, Handa, Tremblay,
  Narang, Van~Wyk, Iqbal, Birchfield, et~al.]{chao2021dexycb}
Yu-Wei Chao, Wei Yang, Yu~Xiang, Pavlo Molchanov, Ankur Handa, Jonathan
  Tremblay, Yashraj~S Narang, Karl Van~Wyk, Umar Iqbal, Stan Birchfield, et~al.
\newblock Dexycb: A benchmark for capturing hand grasping of objects.
\newblock In \emph{CVPR}, 2021.

\bibitem[Chen et~al.(2021{\natexlab{a}})Chen, Chen, Yang, Wu, Li, Xia, and
  Tan]{chen2021i2uv}
Ping Chen, Yujin Chen, Dong Yang, Fangyin Wu, Qin Li, Qingpei Xia, and Yong
  Tan.
\newblock I2uv-handnet: Image-to-uv prediction network for accurate and
  high-fidelity 3d hand mesh modeling.
\newblock In \emph{ICCV}, 2021{\natexlab{a}}.

\bibitem[Chen et~al.(2021{\natexlab{b}})Chen, Tu, Kang, Bao, Zhang, Zhe, Chen,
  and Yuan]{chen2021model}
Yujin Chen, Zhigang Tu, Di~Kang, Linchao Bao, Ying Zhang, Xuefei Zhe, Ruizhi
  Chen, and Junsong Yuan.
\newblock Model-based 3d hand reconstruction via self-supervised learning.
\newblock In \emph{CVPR}, 2021{\natexlab{b}}.

\bibitem[Choi et~al.(2020)Choi, Moon, and Lee]{choi2020pose2mesh}
Hongsuk Choi, Gyeongsik Moon, and Kyoung~Mu Lee.
\newblock Pose2mesh: Graph convolutional network for 3d human pose and mesh
  recovery from a 2d human pose.
\newblock In \emph{ECCV}, 2020.

\bibitem[Christen et~al.(2021)Christen, Kocabas, Aksan, Hwangbo, Song, and
  Hilliges]{christen2021d}
Sammy Christen, Muhammed Kocabas, Emre Aksan, Jemin Hwangbo, Jie Song, and
  Otmar Hilliges.
\newblock D-grasp: Physically plausible dynamic grasp synthesis for hand-object
  interactions.
\newblock \emph{arXiv preprint arXiv:2112.03028}, 2021.

\bibitem[Christen et~al.(2022)Christen, Kocabas, Aksan, Hwangbo, Song, and
  Hilliges]{christen2022d}
Sammy Christen, Muhammed Kocabas, Emre Aksan, Jemin Hwangbo, Jie Song, and
  Otmar Hilliges.
\newblock D-grasp: Physically plausible dynamic grasp synthesis for hand-object
  interactions.
\newblock In \emph{CVPR}, 2022.

\bibitem[de~La~Gorce et~al.(2008)de~La~Gorce, Paragios, and Fleet]{de2008model}
Martin de~La~Gorce, Nikos Paragios, and David~J Fleet.
\newblock Model-based hand tracking with texture, shading and self-occlusions.
\newblock In \emph{CVPR}, 2008.

\bibitem[de~La~Gorce et~al.(2011)de~La~Gorce, Fleet, and Paragios]{de2011model}
Martin de~La~Gorce, David~J Fleet, and Nikos Paragios.
\newblock Model-based 3d hand pose estimation from monocular video.
\newblock \emph{TPAMI}, 33\penalty0 (9):\penalty0 1793--1805, 2011.

\bibitem[Deng et~al.(2018)Deng, Cheng, Xue, Zhou, and Zafeiriou]{deng2018uv}
Jiankang Deng, Shiyang Cheng, Niannan Xue, Yuxiang Zhou, and Stefanos
  Zafeiriou.
\newblock Uv-gan: Adversarial facial uv map completion for pose-invariant face
  recognition.
\newblock In \emph{CVPR}, 2018.

\bibitem[Engelmann et~al.(2017)Engelmann, Kontogianni, Hermans, and
  Leibe]{engelmann2017exploring}
Francis Engelmann, Theodora Kontogianni, Alexander Hermans, and Bastian Leibe.
\newblock Exploring spatial context for 3d semantic segmentation of point
  clouds.
\newblock In \emph{ICCVW}, 2017.

\bibitem[Fan et~al.(2017)Fan, Su, and Guibas]{fan2017point}
Haoqiang Fan, Hao Su, and Leonidas~J Guibas.
\newblock A point set generation network for 3d object reconstruction from a
  single image.
\newblock In \emph{CVPR}, 2017.

\bibitem[Feng et~al.(2018)Feng, Wu, Shao, Wang, and Zhou]{feng2018joint}
Yao Feng, Fan Wu, Xiaohu Shao, Yanfeng Wang, and Xi~Zhou.
\newblock Joint 3d face reconstruction and dense alignment with position map
  regression network.
\newblock In \emph{ECCV}, 2018.

\bibitem[Ge et~al.(2019)Ge, Ren, Li, Xue, Wang, Cai, and Yuan]{ge20193d}
Liuhao Ge, Zhou Ren, Yuncheng Li, Zehao Xue, Yingying Wang, Jianfei Cai, and
  Junsong Yuan.
\newblock 3d hand shape and pose estimation from a single rgb image.
\newblock In \emph{CVPR}, 2019.

\bibitem[Groueix et~al.(2018)Groueix, Fisher, Kim, Russell, and
  Aubry]{groueix2018papier}
Thibault Groueix, Matthew Fisher, Vladimir~G Kim, Bryan~C Russell, and Mathieu
  Aubry.
\newblock A papier-m{\^a}ch{\'e} approach to learning 3d surface generation.
\newblock In \emph{CVPR}, 2018.

\bibitem[G{\"u}ler et~al.(2018)G{\"u}ler, Neverova, and
  Kokkinos]{guler2018densepose}
R{\i}za~Alp G{\"u}ler, Natalia Neverova, and Iasonas Kokkinos.
\newblock Densepose: Dense human pose estimation in the wild.
\newblock In \emph{CVPR}, 2018.

\bibitem[Hampali et~al.(2020)Hampali, Rad, Oberweger, and
  Lepetit]{hampali2020honnotate}
Shreyas Hampali, Mahdi Rad, Markus Oberweger, and Vincent Lepetit.
\newblock Honnotate: A method for 3d annotation of hand and object poses.
\newblock In \emph{CVPR}, 2020.

\bibitem[Hampali et~al.(2021)Hampali, Sarkar, and Lepetit]{hampali2021ho}
Shreyas Hampali, Sayan~Deb Sarkar, and Vincent Lepetit.
\newblock Ho-3d\_v3: Improving the accuracy of hand-object annotations of the
  ho-3d dataset.
\newblock \emph{arXiv preprint arXiv:2107.00887}, 2021.

\bibitem[Hasson et~al.(2019)Hasson, Varol, Tzionas, Kalevatykh, Black, Laptev,
  and Schmid]{hasson2019learning}
Yana Hasson, Gul Varol, Dimitrios Tzionas, Igor Kalevatykh, Michael~J Black,
  Ivan Laptev, and Cordelia Schmid.
\newblock Learning joint reconstruction of hands and manipulated objects.
\newblock In \emph{CVPR}, 2019.

\bibitem[Hasson et~al.(2020)Hasson, Tekin, Bogo, Laptev, Pollefeys, and
  Schmid]{hasson2020leveraging}
Yana Hasson, Bugra Tekin, Federica Bogo, Ivan Laptev, Marc Pollefeys, and
  Cordelia Schmid.
\newblock Leveraging photometric consistency over time for sparsely supervised
  hand-object reconstruction.
\newblock In \emph{CVPR}, 2020.

\bibitem[Hasson et~al.(2021)Hasson, Varol, Schmid, and
  Laptev]{hasson2021towards}
Yana Hasson, G{\"u}l Varol, Cordelia Schmid, and Ivan Laptev.
\newblock Towards unconstrained joint hand-object reconstruction from rgb
  videos.
\newblock In \emph{3DV}, 2021.

\bibitem[He et~al.(2016)He, Zhang, Ren, and Sun]{he2016deep}
Kaiming He, Xiangyu Zhang, Shaoqing Ren, and Jian Sun.
\newblock Deep residual learning for image recognition.
\newblock In \emph{CVPR}, 2016.

\bibitem[Huang et~al.(2020)Huang, Tan, Liu, and Yuan]{huang2020hand}
Lin Huang, Jianchao Tan, Ji~Liu, and Junsong Yuan.
\newblock Hand-transformer: non-autoregressive structured modeling for 3d hand
  pose estimation.
\newblock In \emph{ECCV}, 2020.

\bibitem[Jain and Learned-Miller(2011)]{jain2011online}
Vidit Jain and Erik Learned-Miller.
\newblock Online domain adaptation of a pre-trained cascade of classifiers.
\newblock In \emph{CVPR}, pages 577--584. IEEE, 2011.

\bibitem[Jiang et~al.(2021)Jiang, Liu, Wang, and Wang]{jiang2021hand}
Hanwen Jiang, Shaowei Liu, Jiashun Wang, and Xiaolong Wang.
\newblock Hand-object contact consistency reasoning for human grasps
  generation.
\newblock \emph{arXiv preprint arXiv:2104.03304}, 2021.

\bibitem[Kang et~al.(2021)Kang, Lee, and Lee]{kang2021competitive}
Jiwoo Kang, Seongmin Lee, and Sanghoon Lee.
\newblock Competitive learning of facial fitting and synthesis using uv energy.
\newblock \emph{IEEE Transactions on Systems, Man, and Cybernetics: Systems},
  2021.

\bibitem[Karunratanakul et~al.(2020)Karunratanakul, Yang, Zhang, Black,
  Muandet, and Tang]{karunratanakul2020grasping}
Korrawe Karunratanakul, Jinlong Yang, Yan Zhang, Michael~J Black, Krikamol
  Muandet, and Siyu Tang.
\newblock Grasping field: Learning implicit representations for human grasps.
\newblock In \emph{3DV}, 2020.

\bibitem[Karunratanakul et~al.(2021)Karunratanakul, Spurr, Fan, Hilliges, and
  Tang]{karunratanakul2021skeleton}
Korrawe Karunratanakul, Adrian Spurr, Zicong Fan, Otmar Hilliges, and Siyu
  Tang.
\newblock A skeleton-driven neural occupancy representation for articulated
  hands.
\newblock In \emph{3DV}, 2021.

\bibitem[Khamis et~al.(2015)Khamis, Taylor, Shotton, Keskin, Izadi, and
  Fitzgibbon]{khamis2015learning}
Sameh Khamis, Jonathan Taylor, Jamie Shotton, Cem Keskin, Shahram Izadi, and
  Andrew Fitzgibbon.
\newblock Learning an efficient model of hand shape variation from depth
  images.
\newblock In \emph{CVPR}, 2015.

\bibitem[Kolotouros et~al.(2019)Kolotouros, Pavlakos, Black, and
  Daniilidis]{kolotouros2019learning}
Nikos Kolotouros, Georgios Pavlakos, Michael~J Black, and Kostas Daniilidis.
\newblock Learning to reconstruct 3d human pose and shape via model-fitting in
  the loop.
\newblock In \emph{ICCV}, 2019.

\bibitem[Kulon et~al.(2020)Kulon, Guler, Kokkinos, Bronstein, and
  Zafeiriou]{kulon2020weakly}
Dominik Kulon, Riza~Alp Guler, Iasonas Kokkinos, Michael~M Bronstein, and
  Stefanos Zafeiriou.
\newblock Weakly-supervised mesh-convolutional hand reconstruction in the wild.
\newblock In \emph{CVPR}, 2020.

\bibitem[Lee and Lee(2020)]{lee2020uncertainty}
Gun-Hee Lee and Seong-Whan Lee.
\newblock Uncertainty-aware mesh decoder for high fidelity 3d face
  reconstruction.
\newblock In \emph{CVPR}, 2020.

\bibitem[Lee et~al.(2020)Lee, Cho, Kim, Inouye, and Kwak]{lee2020styleuv}
Myunggi Lee, Wonwoong Cho, Moonheum Kim, David Inouye, and Nojun Kwak.
\newblock Styleuv: Diverse and high-fidelity uv map generative model.
\newblock \emph{arXiv preprint arXiv:2011.12893}, 2020.

\bibitem[Li et~al.(2021)Li, Yang, Zhan, Lv, Xu, Li, and Lu]{li2021artiboost}
Kailin Li, Lixin Yang, Xinyu Zhan, Jun Lv, Wenqiang Xu, Jiefeng Li, and Cewu
  Lu.
\newblock Artiboost: Boosting articulated 3d hand-object pose estimation via
  online exploration and synthesis.
\newblock \emph{arXiv preprint arXiv:2109.05488}, 2021.

\bibitem[Lin et~al.(2021{\natexlab{a}})Lin, Wang, and Liu]{lin2021end}
Kevin Lin, Lijuan Wang, and Zicheng Liu.
\newblock End-to-end human pose and mesh reconstruction with transformers.
\newblock In \emph{CVPR}, 2021{\natexlab{a}}.

\bibitem[Lin et~al.(2021{\natexlab{b}})Lin, Wang, and Liu]{lin2021mesh}
Kevin Lin, Lijuan Wang, and Zicheng Liu.
\newblock Mesh graphormer.
\newblock In \emph{ICCV}, 2021{\natexlab{b}}.

\bibitem[Liu et~al.(2017)Liu, Li, Zhang, Zhou, Ye, Wang, and Lu]{liu20173dcnn}
Fangyu Liu, Shuaipeng Li, Liqiang Zhang, Chenghu Zhou, Rongtian Ye, Yuebin
  Wang, and Jiwen Lu.
\newblock 3dcnn-dqn-rnn: A deep reinforcement learning framework for semantic
  parsing of large-scale 3d point clouds.
\newblock In \emph{ICCV}, 2017.

\bibitem[Loper et~al.(2015)Loper, Mahmood, Romero, Pons-Moll, and
  Black]{loper2015smpl}
Matthew Loper, Naureen Mahmood, Javier Romero, Gerard Pons-Moll, and Michael~J
  Black.
\newblock Smpl: A skinned multi-person linear model.
\newblock \emph{ACM transactions on graphics (TOG)}, 34\penalty0 (6):\penalty0
  1--16, 2015.

\bibitem[Melax et~al.(2013)Melax, Keselman, and Orsten]{melax2013dynamics}
Stan Melax, Leonid Keselman, and Sterling Orsten.
\newblock Dynamics based 3d skeletal hand tracking.
\newblock In \emph{Proceedings of the ACM SIGGRAPH Symposium on Interactive 3D
  Graphics and Games}, pages 184--184, 2013.

\bibitem[Moon and Lee(2020)]{moon2020i2l}
Gyeongsik Moon and Kyoung~Mu Lee.
\newblock I2l-meshnet: Image-to-lixel prediction network for accurate 3d human
  pose and mesh estimation from a single rgb image.
\newblock In \emph{ECCV}, 2020.

\bibitem[Mullapudi et~al.(2019)Mullapudi, Chen, Zhang, Ramanan, and
  Fatahalian]{mullapudi2019online}
Ravi~Teja Mullapudi, Steven Chen, Keyi Zhang, Deva Ramanan, and Kayvon
  Fatahalian.
\newblock Online model distillation for efficient video inference.
\newblock In \emph{ICCV}, 2019.

\bibitem[Oikonomidis et~al.(2011)Oikonomidis, Kyriazis, and
  Argyros]{oikonomidis2011efficient}
Iason Oikonomidis, Nikolaos Kyriazis, and Antonis~A Argyros.
\newblock Efficient model-based 3d tracking of hand articulations using kinect.
\newblock In \emph{BmVC}, 2011.

\bibitem[Qi et~al.(2017{\natexlab{a}})Qi, Su, Mo, and Guibas]{qi2017pointnet}
Charles~R Qi, Hao Su, Kaichun Mo, and Leonidas~J Guibas.
\newblock Pointnet: Deep learning on point sets for 3d classification and
  segmentation.
\newblock In \emph{CVPR}, 2017{\natexlab{a}}.

\bibitem[Qi et~al.(2017{\natexlab{b}})Qi, Yi, Su, and Guibas]{qi2017pointnet++}
Charles~Ruizhongtai Qi, Li~Yi, Hao Su, and Leonidas~J Guibas.
\newblock Pointnet++: Deep hierarchical feature learning on point sets in a
  metric space.
\newblock \emph{NeurIPS}, 2017{\natexlab{b}}.

\bibitem[Qian et~al.(2020)Qian, Wang, Mueller, Bernard, Golyanik, and
  Theobalt]{qian2020parametric}
Neng Qian, Jiayi Wang, Franziska Mueller, Florian Bernard, Vladislav Golyanik,
  and Christian Theobalt.
\newblock Parametric hand texture model for 3d hand reconstruction and
  personalization.
\newblock In \emph{ECCV}, 2020.

\bibitem[Ranjan et~al.(2018)Ranjan, Bolkart, Sanyal, and
  Black]{ranjan2018generating}
Anurag Ranjan, Timo Bolkart, Soubhik Sanyal, and Michael~J Black.
\newblock Generating 3d faces using convolutional mesh autoencoders.
\newblock In \emph{ECCV}, 2018.

\bibitem[Ravi et~al.(2020)Ravi, Reizenstein, Novotny, Gordon, Lo, Johnson, and
  Gkioxari]{ravi2020pytorch3d}
Nikhila Ravi, Jeremy Reizenstein, David Novotny, Taylor Gordon, Wan-Yen Lo,
  Justin Johnson, and Georgia Gkioxari.
\newblock Accelerating 3d deep learning with pytorch3d.
\newblock \emph{arXiv:2007.08501}, 2020.

\bibitem[Romero et~al.(2017{\natexlab{a}})Romero, Tzionas, and
  Black]{MANO:SIGGRAPHASIA:2017}
Javier Romero, Dimitrios Tzionas, and Michael~J. Black.
\newblock Embodied hands: Modeling and capturing hands and bodies together.
\newblock \emph{ACM Transactions on Graphics, (Proc. SIGGRAPH Asia)},
  36\penalty0 (6), November 2017{\natexlab{a}}.

\bibitem[Romero et~al.(2017{\natexlab{b}})Romero, Tzionas, and
  Black]{romero2017embodied}
Javier Romero, Dimitrios Tzionas, and Michael~J Black.
\newblock Embodied hands: Modeling and capturing hands and bodies together.
\newblock \emph{ACM Transactions on Graphics (ToG)}, 2017{\natexlab{b}}.

\bibitem[Ronneberger et~al.(2015)Ronneberger, Fischer, and
  Brox]{ronneberger2015u}
Olaf Ronneberger, Philipp Fischer, and Thomas Brox.
\newblock U-net: Convolutional networks for biomedical image segmentation.
\newblock In \emph{International Conference on Medical image computing and
  computer-assisted intervention}, 2015.

\bibitem[Sohn et~al.(2015)Sohn, Lee, and Yan]{cvae}
Kihyuk Sohn, Honglak Lee, and Xinchen Yan.
\newblock Learning structured output representation using deep conditional
  generative models.
\newblock \emph{NeurIPS}, 2015.

\bibitem[Spurr et~al.(2020)Spurr, Iqbal, Molchanov, Hilliges, and
  Kautz]{spurr2020weakly}
Adrian Spurr, Umar Iqbal, Pavlo Molchanov, Otmar Hilliges, and Jan Kautz.
\newblock Weakly supervised 3d hand pose estimation via biomechanical
  constraints.
\newblock In \emph{ECCV}, 2020.

\bibitem[Sun et~al.(2020)Sun, Wang, Liu, Miller, Efros, and Hardt]{sun2020test}
Yu~Sun, Xiaolong Wang, Zhuang Liu, John Miller, Alexei Efros, and Moritz Hardt.
\newblock Test-time training with self-supervision for generalization under
  distribution shifts.
\newblock In \emph{ICML}, pages 9229--9248, 2020.

\bibitem[Taheri et~al.(2020)Taheri, Ghorbani, Black, and
  Tzionas]{taheri2020grab}
Omid Taheri, Nima Ghorbani, Michael~J Black, and Dimitrios Tzionas.
\newblock Grab: A dataset of whole-body human grasping of objects.
\newblock In \emph{ECCV}, pages 581--600. Springer, 2020.

\bibitem[Tzionas et~al.(2016)Tzionas, Ballan, Srikantha, Aponte, Pollefeys, and
  Gall]{tzionas2016capturing}
Dimitrios Tzionas, Luca Ballan, Abhilash Srikantha, Pablo Aponte, Marc
  Pollefeys, and Juergen Gall.
\newblock Capturing hands in action using discriminative salient points and
  physics simulation.
\newblock \emph{IJCV}, 118\penalty0 (2):\penalty0 172--193, 2016.

\bibitem[Wan et~al.(2020)Wan, Probst, Van~Gool, and Yao]{wan2020dual}
Chengde Wan, Thomas Probst, Luc Van~Gool, and Angela Yao.
\newblock Dual grid net: Hand mesh vertex regression from single depth maps.
\newblock In \emph{ECCV}, 2020.

\bibitem[Xiang et~al.(2017)Xiang, Schmidt, Narayanan, and
  Fox]{xiang2017posecnn}
Yu~Xiang, Tanner Schmidt, Venkatraman Narayanan, and Dieter Fox.
\newblock Posecnn: A convolutional neural network for 6d object pose estimation
  in cluttered scenes.
\newblock \emph{arXiv preprint arXiv:1711.00199}, 2017.

\bibitem[Yang and Yao(2019)]{yang2019disentangling}
Linlin Yang and Angela Yao.
\newblock Disentangling latent hands for image synthesis and pose estimation.
\newblock In \emph{CVPR}, 2019.

\bibitem[Yang et~al.(2021)Yang, Zhan, Li, Xu, Li, and Lu]{yang2021cpf}
Lixin Yang, Xinyu Zhan, Kailin Li, Wenqiang Xu, Jiefeng Li, and Cewu Lu.
\newblock Cpf: Learning a contact potential field to model the hand-object
  interaction.
\newblock In \emph{ICCV}, 2021.

\bibitem[Yang et~al.(2018)Yang, Feng, Shen, and Tian]{yang2018foldingnet}
Yaoqing Yang, Chen Feng, Yiru Shen, and Dong Tian.
\newblock Foldingnet: Point cloud auto-encoder via deep grid deformation.
\newblock In \emph{CVPR}, pages 206--215, 2018.

\bibitem[Yao et~al.(2019)Yao, Fang, Wu, Feng, and Li]{yao2019densebody}
Pengfei Yao, Zheng Fang, Fan Wu, Yao Feng, and Jiwei Li.
\newblock Densebody: Directly regressing dense 3d human pose and shape from a
  single color image.
\newblock \emph{arXiv preprint arXiv:1903.10153}, 2019.

\bibitem[Zeng et~al.(2020)Zeng, Ouyang, Luo, Liu, and Wang]{zeng20203d}
Wang Zeng, Wanli Ouyang, Ping Luo, Wentao Liu, and Xiaogang Wang.
\newblock 3d human mesh regression with dense correspondence.
\newblock In \emph{CVPR}, 2020.

\bibitem[Zhang et~al.(2020)Zhang, Fang, Wah, and Torr]{zhang2020deep}
Feihu Zhang, Jin Fang, Benjamin Wah, and Philip Torr.
\newblock Deep fusionnet for point cloud semantic segmentation.
\newblock In \emph{ECCV}, 2020.

\bibitem[Zhu et~al.(2019)Zhu, Zuo, Wang, Cao, and Yang]{zhu2019detailed}
Hao Zhu, Xinxin Zuo, Sen Wang, Xun Cao, and Ruigang Yang.
\newblock Detailed human shape estimation from a single image by hierarchical
  mesh deformation.
\newblock In \emph{CVPR}, 2019.

\bibitem[Zimmermann and Brox(2017)]{zimmermann2017learning}
Christian Zimmermann and Thomas Brox.
\newblock Learning to estimate 3d hand pose from single rgb images.
\newblock In \emph{ICCV}, 2017.

\bibitem[Zimmermann et~al.(2019)Zimmermann, Ceylan, Yang, Russell, Argus, and
  Brox]{zimmermann2019freihand}
Christian Zimmermann, Duygu Ceylan, Jimei Yang, Bryan Russell, Max Argus, and
  Thomas Brox.
\newblock Freihand: A dataset for markerless capture of hand pose and shape
  from single rgb images.
\newblock In \emph{ICCV}, 2019.

\end{thebibliography}
\end{document}